\documentclass{bmvc2k}
\usepackage{multirow}
\usepackage{amssymb}
\usepackage{marvosym}
\usepackage[marginal]{footmisc}

\title{Cross-Modal Scene Semantic Alignment for Image Complexity Assessment}

\addauthor{Yuqing Luo}{27620195575393@stu.xmu.edu.cn}{1}
\addauthor{Yixiao Li\textsuperscript{*}}{18335310648@163.com}{2}
\addauthor{Jiang Liu}{liuj137@cardiff.ac.uk}{1}
\addauthor{Jun Fu}{fujun@mail.ustc.edu.cn}{1}
\addauthor{Hadi Amirpour}{Hadi.Amirpour@aau.at}{3}
\addauthor{Guanghui Yue}{yueguanghui@szu.edu.cn}{4}
\addauthor{Baoquan Zhao}{zhaobaoquan@mail.sysu.edu.cn}{5}
\addauthor{Padraig Corcoran}{corcoranp@cardiff.ac.uk}{1}
\addauthor{Hantao Liu}{liuh35@cardiff.ac.uk}{1}
\addauthor{Wei Zhou\textsuperscript{\Letter}}{zhouw26@cardiff.ac.uk}{1}
\addinstitution{
 School of Computer Science\\
 Cardiff University\\
 United Kingdom
}
\addinstitution{
 School of Mathematical Sciences\\
 Beihang University\\
 China
}
\addinstitution{
Department of Information Technology\\
University of Klagenfurt\\
Austria
}
\addinstitution{
School of Biomedical Engineering\\
Shenzhen University\\
China
}
\addinstitution{
School of Artificial Intelligence\\
Sun Yat-sen University\\
China
}
\runninghead{Yuqing Luo, Yixiao Li, et al.}{CM-SSA}

\begin{document}

\maketitle
\footnote{\noindent \textsuperscript{*} denotes the co-first author. \textsuperscript{\Letter} denotes the corresponding author.}
\begin{abstract}
Image complexity assessment (ICA) is a challenging task in perceptual evaluation due to the subjective nature of human perception and the inherent semantic diversity in real-world images. Existing ICA methods predominantly rely on hand-crafted or shallow convolutional neural network-based features of a single visual modality, which are insufficient to fully capture the perceived representations closely related to image complexity. Recently, cross-modal scene semantic information has been shown to play a crucial role in various computer vision tasks, particularly those involving perceptual understanding. However, the exploration of cross-modal scene semantic information in the context of ICA remains unaddressed. Therefore, in this paper, we propose a novel ICA method called Cross-Modal Scene Semantic Alignment (CM-SSA), which leverages scene semantic alignment from a cross-modal perspective to enhance ICA performance, enabling complexity predictions to be more consistent with subjective human perception. Specifically, the proposed CM-SSA consists of a complexity regression branch and a scene semantic alignment branch. The complexity regression branch estimates image complexity levels under the guidance of the scene semantic alignment branch, while the scene semantic alignment branch is used to align images with corresponding text prompts that convey rich scene semantic information by pair-wise learning. Extensive experiments on several ICA datasets demonstrate that the proposed CM-SSA significantly outperforms state-of-the-art approaches. Codes are available at \url{https://github.com/XQ2K/First-Cross-Model-ICA}.
\end{abstract}

\section{Introduction}
\label{sec:intro}
The study of image complexity is particularly relevant to psychology and computer vision research~\cite{WOS:000340226200024,article}. In psychology, image complexity plays a critical role in shaping visual aesthetics and influencing emotional responses~\cite{McCormack_Lomas_2020,TUCH2009703}, while in computer vision, it serves as a fundamental attribute across various tasks. Automating image complexity assessment (ICA) has been proven beneficial for applications that include fire detection~\cite{li2021evaluation}, image steganography~\cite{Grover_Yadav_Chauhan_Kamya_2018}, visual quality assessment~\cite{zhou2025perceptual}, aesthetic image classification~\cite{romero2012using}, and image enhancement~\cite{Cheng2016DigitalIE}. However, the definition of image complexity remains a complex challenge. Subjectively, it represents the cognitive challenge that a human observer faces in interpreting or describing an image, factoring in global patterns and localized details, such as textures or fine structures~\cite{inproceedings,Forsythe_2008}. Objectively, image complexity encompasses the amount of detail, diversity, and structural variety~\cite{WOS:000285044800192}.

\begin{figure*}[t]
\centerline{\includegraphics[width=0.95\linewidth]{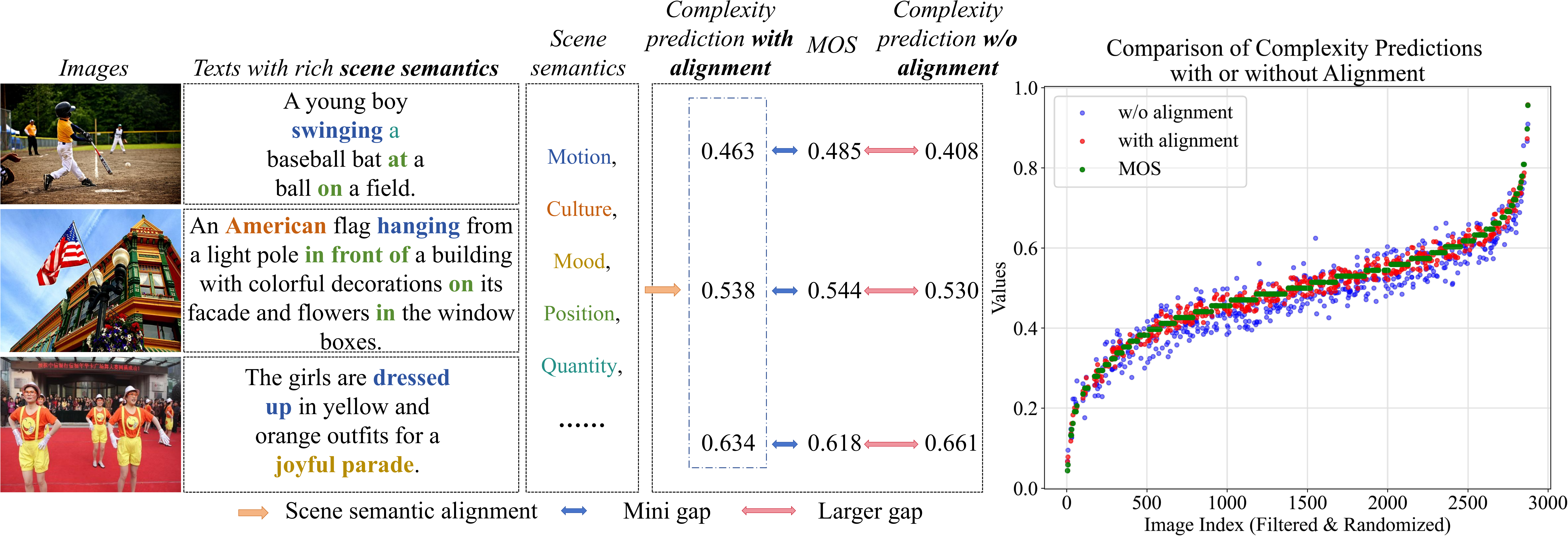}}
\caption{Examples of image complexity predictions with and without the guidance of our proposed cross-modal scene semantic alignment. Here, we adopt scene semantic information such as motion, culture, mood, spatial position, and quantity, which capture the relationships among the key visual entities (e.g., people, objects) within an image. The scatter plot visualizes the predictions of complexity scores on the IC9600 test set.}
\label{fig1}
\end{figure*}

In the literature, existing ICA methods can be broadly categorized into hand-crafted feature-based and deep learning-based methods. Among hand-crafted feature-based ICA approaches, early studies conceptualized complexity through statistical analysis of certain attributes of images. For example, multidimensional attributes such as object count, clutter, openness, symmetry, and color variety~\cite{WOS:000285044800192} are considered. Subsequent works employed machine learning models like gradient-boosted trees to regress complexity features~\cite{Sun_Yamasaki_Aizawa_2014}, and aligned computational measures with subjective human perception~\cite{inproceedings}. Other studies explored spatial information and compression-based metrics~\cite{6603194}, eye movement analysis~\cite{6116371}, detailed feature designs capturing global, local, as well as salient characteristics~\cite{GUO2018110}, compression errors~\cite{Machado_Romero_Nadal_Santos_Correia_Carballal_2015}, and entropy~\cite{Stamps_2002,Rosenholtz_Li_Nakano_2007}. However, these methods have limited generalization for the ICA task compared to deep learning-based approaches.

In recent years, deep learning-based ICA methods have leveraged neural networks to capture richer representations. Early works used the convolutional neural network (CNN) to extract features from texture and salient regions for complexity prediction~\cite{7340938,8936095, SARAEE2020102949,guo2023image}. However, the limited dataset size constrained the evaluation of their effectiveness and generalization ability. To address this, the IC9600 dataset~\cite{9999482} was introduced, providing 9,600 labeled images along with a dual-branch CNN model, i.e., ICNet. To further reduce the annotation costs of constructing a large-scale dataset, Liu et al. \cite{liu2024contrastive} employed unsupervised contrastive learning based on MoCo v2 \cite{chen2020improved} for complexity representation learning. And attention-based models are further explored~\cite{celona2024use,liu2025clicv2,li2025unlocking}.

Hayes et al.~\cite{hayes2019scene} demonstrated through subjective perceptual experiments that scene semantic information involuntarily guides visual attention during search tasks. Thus, such information is closely linked to the subjective perception of images. Despite the above-mentioned advancements, existing ICA models, whether based on hand-crafted or CNN-derived features, have largely overlooked the exploration of scene semantic information. In particular, aspects such as object motion, cultural context, mood, spatial position, and quantity—factors that critically reflect the interrelationships among image key visual entities, as illustrated in Fig.~\ref{fig1}—have not been effectively integrated into the current ICA modeling frameworks.

However, the significance of scene semantic information has been extensively demonstrated in various computer vision tasks. Specifically, Delaitre et al.~\cite{delaitre2012scene} leveraged scene semantics by incorporating the distributional patterns of object-related super-pixels over time, capturing how people interact with objects, thereby improving pose estimation and semantic labeling tasks. Wu et al.~\cite{wu2016harnessing} incorporated object and scene semantics, extracted via pre-trained detectors, into the video understanding pipeline, subsequently fusing these semantic cues to improve large-scale video understanding. Wei et al.~\cite{wei2021integrating} integrated textual scene semantic topic words into image captioning to enable the model to generate more accurate and scene-specific captions. Siahaan et al. \cite{siahaan2018semantic} improved no-reference image quality assessment by using image semantic information (i.e., scene and object categories).

Inspired by the crucial role of scene semantic information in many other visual tasks, we aim to enhance the ICA task by explicitly exploring such information. Although such semantics could theoretically be extracted through pre-trained tasks like image classification or semantic segmentation, the ICA datasets~\cite{9999482,akagunduz2019defining,SaraeeJaBe20} often lack the corresponding ground truth labels for these specific tasks. Thus, the absence of explicit semantic labels makes it challenging to directly incorporate scene semantic information into the complexity assessment process.

Since capturing scene semantic information from the vision-language framework is an efficient way~\cite{wei2021integrating}. To address the above gap and enhance the model’s focus on scene semantic information, we propose a new ICA method
called Cross-Modal Scene Semantic Alignment (CM-SSA). Our proposed CM-SSA adopts a dual-branch framework that refines visual features through image-text alignment guided by scene semantics. The complexity regression branch predicts complexity levels directly using image-prompt pair-wise learning based on complexity categories such as ``high complexity'' and ``moderate complexity''. Meanwhile, the scene semantic alignment branch refines image features by aligning them with scene descriptions automatically generated by instruction-tuned vision-language models such as InstructBLIP~\cite{WOS:001224281507038}. As shown in Fig.~\ref{fig1}, incorporating the proposed cross-modal scene semantic alignment leads to complexity predictions that are more consistent with human perception.

In summary, the main contributions of this paper are as follows:
\begin{enumerate}
    \item We propose the first cross-modal metric, namely Cross-Modal Scene Semantic Alignment (CM-SSA), for image complexity assessment.

    \item We introduce a dual-branch framework, including the complexity regression branch and the scene semantic alignment branch, for exploiting cross-modal scene semantic information to assist the complexity prediction.

    \item Experiments on several ICA datasets, including IC9600, VISC-C, and SAVOIAS, demonstrate that our proposed CM-SSA outperforms state-of-the-art methods. Ablation studies further validate the effectiveness of each proposed key component.
\end{enumerate}

\begin{figure*}[t]
\centerline{\includegraphics[width=0.9\linewidth]{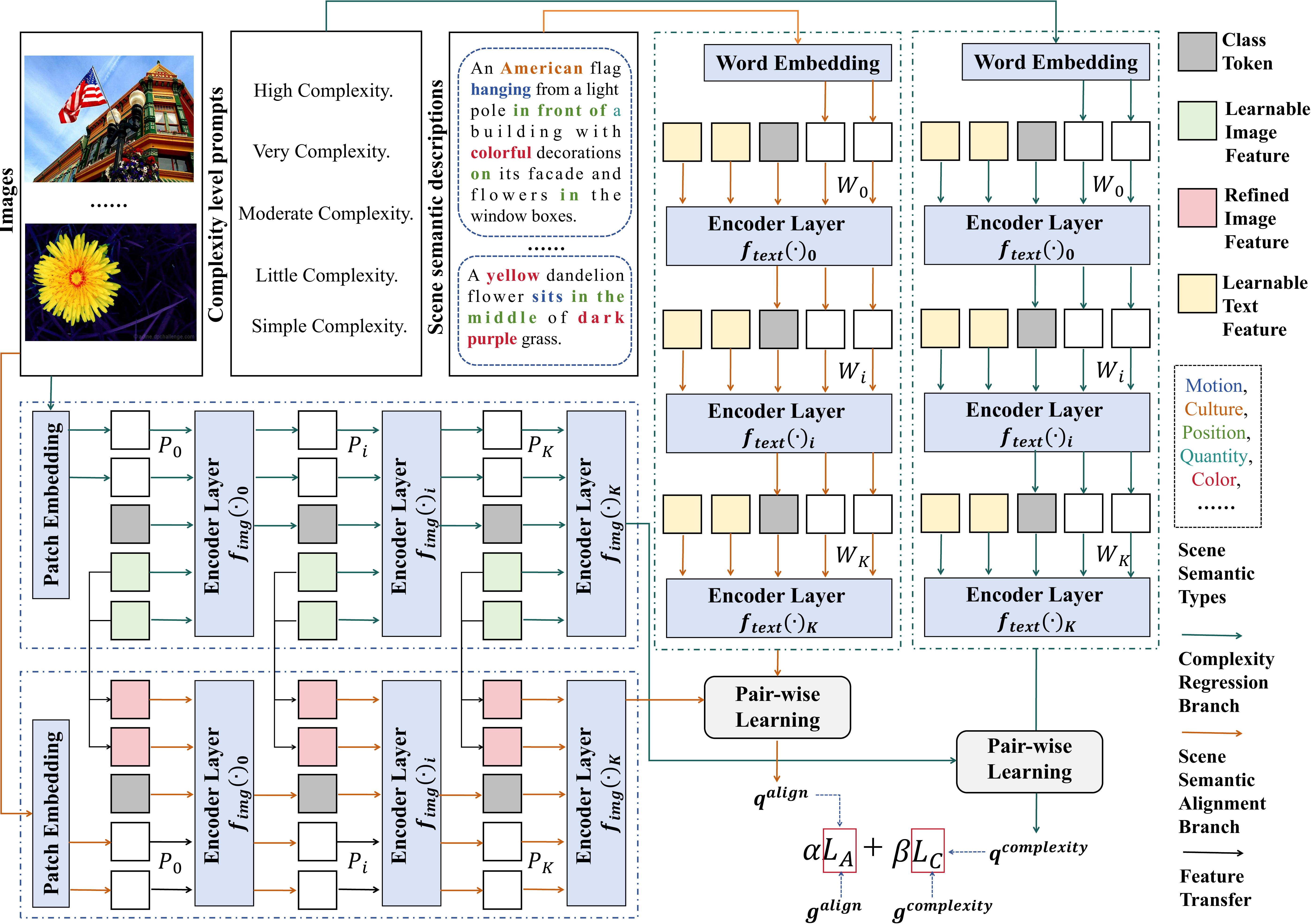}}
\caption{The framework of the proposed \textbf{ Cross-Modal Scene Semantic Alignment (CM-SSA)}. The hyperparameters $\alpha$ and $\beta$ control the relative weights of the \textbf{complexity regression branch} and \textbf{scene semantic alignment branch}. The alignment loss $L_A$ encourages the image features to align with the scene semantic text features, using a ground truth of a vector of all ones (i.e., $g^{align}$) to ensure maximum alignment. The complexity loss $L_C$ predicts image complexity levels, where the ground truth $g^{complexity}$ corresponds to the MOS (Mean Opinion Score) of image complexity.}
\label{fig2}
\end{figure*}

\section{Related Work}
Existing ICA methods can be broadly categorized into hand-crafted feature-based and deep learning-based approaches. Among the hand-crafted approaches, Oliva et al.~\cite{WOS:000285044800192} conceptualized visual complexity as a multidimensional representation involving factors such as the number of objects, clutter, openness, symmetry, organization, and variety of colors. Sun et al.~\cite{Sun_Yamasaki_Aizawa_2014} adopted gradient-boosted trees to regress image complexity features like composition, statistical properties, and distribution. Purchase et al.~\cite{inproceedings} quantified visual complexity by aligning it with participants' subjective perception, concluding that while the subjective concept of ``complexity'' is consistent across individuals and groups, it is not easily correlated with straightforward computational metrics. Yu et al.~\cite{6603194} found a strong association between spatial information metrics and compression-based complexity measures. Silva et al.~\cite{6116371} investigated the relationship between human eye movements and various computational attention models in the context of ICA. Guo et al.~\cite{GUO2018110} considered how global and local characteristics in paintings influence overall impressions and detailed perceptions. Penousal et al.~\cite{Machado_Romero_Nadal_Santos_Correia_Carballal_2015} used edge detection techniques, as well as metrics based on image compression errors, to estimate visual complexity. Additionally, entropy-based ICA methods have also been actively explored. Stamps et al.~\cite{Stamps_2002} investigated the relationship between visual diversity and statistical entropy. Ruth et al.~\cite{Rosenholtz_Li_Nakano_2007} proposed three visual clutter metrics (feature congestion, subband entropy, and edge density), which improved the generalization of visual search models and highlighted the crucial role of color diversity in visual clutter estimation. However, these methods rely on predefined features and prior assumptions, limiting their adaptability and semantic understanding.

Recently, deep learning-based ICA methods have exploited neural networks for richer representations. For example, Chen et al.~\cite{7340938} used networks to analyze texture and salient regions, while Abdelwahab et al.~\cite{8936095} and Saraee et al.~\cite{SARAEE2020102949} employed the CNN with SVM or ridge regression to predict the visual complexity. Ivanovici et al.~\cite{9301137} explored the CNN for modeling fractal image complexity using modified ResNet-18. However, the effectiveness and generalization of these models are limited due to the lack of large-scale ICA datasets. Therefore, the IC9600 dataset~\cite{9999482} was introduced as the first large-scale ICA dataset, containing 9,600 images. To avoid the high annotation cost of constructing the large-scale IC9600 dataset, Liu et al.~\cite{liu2024contrastive} then utilized unsupervised learning based on the MoCo v2~\cite{chen2020improved} for the representation learning of image complexity. However, these methods often use shallow networks and single-image modality as input, failing to capture high-level semantics and multi-modal cues.

\section{Proposed CM-SSA Method}
In this section, we introduce our proposed ICA metric, namely the Cross-modal Scene Semantic Alignment (CM-SSA), designed to guide the image encoder in extracting more complexity-related scene semantic information through cross-modal alignment between images and corresponding scene semantic descriptions. Fig. \ref{fig2} shows the overall framework of the proposed CM-SSA, which consists of two branches: the complexity regression branch and the scene semantic alignment branch. Both branches utilize a cross-modal pair-wise learning between images and texts to fine-tune the contrastive language-image pretraining (CLIP) model. We begin by briefly introducing the image and text encoders of the CLIP model, followed by a detailed explanation of the proposed CM-SSA framework.

\subsection{CLIP Model}
The CLIP model~\cite{radford2021learning} maps images and text into a shared high-dimensional embedding space for cross-modal alignment. The image encoder, typically based on a Vision Transformer~\cite{dosovitskiy2020image} or ResNet~\cite{he2016deep}, processes an input image $X^{\text{Img}}$ by tokenizing it into $M$ patches. Each patch is embedded into a $d_v$-dimensional latent vector, resulting in a patch matrix $P_0 \in \mathbb{R}^{M \times d_v}$. A learnable class token $c_0 \in \mathbb{R}^{d_v}$ is then added to $P_0$, and the sequence $[c_0, P_0]$ is fed into the initial transformer block ${f_{\text{img}(\cdot)}}_{0}$. The process is recursively repeated $K-1$ times across subsequent blocks:
\begin{equation}
\small
[c_i, P_i] = {f_{\text{img}}([c_{i-1}, P_{i-1}])}_{i}, \quad i = 1, \ldots, K.
\end{equation}
The final image representation $Z^{\text{Img}} \in \mathbb{R}^{D}$ is obtained by projecting the last class token $c_K$ into a $D$-dimensional latent space. $f_{\text{img}}(\cdot)$ denotes the overall image encoder. Similarly, the text encoder $f_{\text{text}}(\cdot)$, based on a transformer architecture, tokenizes and embeds the input textual prompt $X^{\text{Text}}$ into a sequence of $L$ words, each mapped into a $d_l$-dimensional embedding. The initial word embedding matrix $W_0 = [w_0^1, \ldots, w_0^L] \in \mathbb{R}^{L \times d_l}$ is then passed through $K$ transformer layers as follows:
\begin{equation}
\small
[W_i] = {f_{\text{text}}(W_{i-1})}_{i}, \quad i = 1, \ldots, K.
\end{equation}
The final text representation $Z^{\text{Text}} \in \mathbb{R}^{D}$ is obtained by projecting the last token $w_K^L$ into the same latent space as the image representation.
\subsection{Cross-Modal Scene Semantic Alignment for ICA}
We aim to align images and their corresponding scene semantic information in a shared embedding space, enabling the rich scene semantics in the text to guide the learning of complexity-related features in the image embedding. A straightforward implementation involves fine-tuning the CLIP model by inputting the image and its corresponding scene semantic textual description to directly regress to complexity levels. 

Since the IC9600 dataset~\cite{9999482} for the ICA task does not contain scene semantic textual descriptions, we first generate such descriptions $X^{\text{Scene}}$ for each input image $X^{\text{Img}}$ using the InstructBLIP~\cite{WOS:001224281507038}. Examples of these descriptions include phrases such as ``An American flag hanging from a light pole in front of a building with colorful decorations on its facade and flowers in the window boxes'', as shown in Fig. \ref{fig2}. 

Subsequently, we input the image-text pair $[X^{\text{Img}}, X^{\text{Scene}}]$ into the CLIP model's image encoder and text encoder, respectively, obtaining the corresponding embeddings $[Z^{\text{Img}}, Z^{\text{Scene}}]$. These embeddings are aligned in a shared high-dimensional embedding space through image-text pair-wise learning. Specifically, let the input image set be $\{X^{\text{Img}}_i, i = 1, 2, \dots, N\}$, where $N$ denotes the number of input images in a batch, and let the corresponding textual description set be $\{X^{\text{Scene}}_i, i = 1, 2, \dots, N\}$. The embeddings obtained through the image and text encoders can be expressed as:
\begin{equation}
\small
  [Z^{\text{Img}}_i, Z^{\text{Scene}}_i] = [f_{\text{img}}(X^{\text{Img}}_i), f_{\text{text}}(X^{\text{Scene}}_i)], \quad i = 1, 2, \dots, N.
\label{eq3}
\end{equation}

Inspired by AGIQA~\cite{fu2024vision}, we utilized additional learnable prompts for better learning capability, as shown in Fig. \ref{fig2}. The cosine similarity between the image and text embeddings is then computed to facilitate alignment:
\begin{equation}
\small
s_{i}^{complexity}=\frac{{Z^{\text{Img}}_{i}} \odot {Z^{\text{Scene}}_{j}}}{\|{Z^{\text{Img}}_{i}}\| \cdot\left\|{Z^{\text{Scene}}_{j}}\right\|}, i,j \in\{1, 2, \dots, N\},
\label{eq4}
\end{equation}
and the result is regressed to predict the complexity score:
\begin{equation}
\small
q_{i}^{complexity}=\frac{e^{s_{i}^{complexity}}}{\sum_{j=1}^{N}e^{s_{j}^{complexity}}}, i\in 1, 2, \dots, N.
\label{eq5}
\end{equation}

However, \textbf{the experimental results of this direct approach reveal a significant gap from the dual-branch framework illustrated in the ablation study in Table \ref{tab:branch_performance}.} We believe this is because scene semantic textual descriptions, while reflecting the scene content of images, do not explicitly align with specific complexity levels. Consequently, aligning scene semantic text with complexity predictions is less effective. To address this, we propose the CM-SSA framework consisting of both a complexity regression branch (i.e., $\rm Branch_C$) and a scene semantic alignment branch (i.e., $\rm Branch_A$).

In the complexity regression branch, textual prompts of complexity levels $X_{k}^{\text{Complexity}}, k=1, 2, 3, 4, 5$ (\textit{i.e.,} Simple Complexity, Little Complexity, Moderate Complexity, Very Complexity, High Complexity) are utilized. Following the calculation processes of Eq. (\ref{eq3}, \ref{eq4}, \ref{eq5}), the complexity predictions are calculated as follows:
\begin{equation}
\small
\begin{aligned}
    &\ [Z^{\text{Img}}_i, Z^{\text{Complexity}}_{k}] = [f_{\text{img}}(X^{\text{Img}}_i), f_{\text{text}}(X^{\text{Complexity}}_k)], \\
    &\ s_{ik}^{complexity}=\frac{{Z^{\text{Img}}_{i}} \odot {Z^{\text{Complexity}}_{k}}}{\|{Z^{\text{Img}}_{i}}\| \cdot\left\|{Z^{\text{Complexity}}_{k}}\right\|}, \\
    &\ q_{i}^{complexity}=\frac{e^{s_{i1}^{complexity}}}{\sum_{k=1}^{5}e^{s_{ik}^{complexity}}}, i \in\{1, 2, \dots, N\}, k \in\{1, 2, 3, 4, 5\}.
\end{aligned}
\end{equation}
The complexity loss is calculated by the MSE loss:
\begin{equation}
\small
L_{C} =\frac{1}{N} \sum_{i=1}^{N}\left\|q^{complexity}_{i}-g^{complexity}_{i}\right\|_{2}^{2},
\end{equation}
where $g^{complexity}$ is the MOS of image complexity.

Furthermore, to involve the scene semantic information in refining the image feature learning, we conceive the scene semantic alignment branch that fully aligns images with corresponding scene semantic textual descriptions. 

The input to this branch is the image-scene semantic description pair $[X^{\text{Img}}_i, X^{\text{Scene}}_i], i = 1, 2, \dots, N$. The cosine similarity and alignment level between the image and scene semantic text are computed as:
\begin{equation}
\small
\begin{aligned}
    &\ s_{i}^{scene}=\frac{{Z^{\text{Img}}_{i}} \odot {Z^{\text{Scene}}_{j}}}{\|{Z^{\text{Img}}_{i}}\| \cdot\left\|{Z^{\text{Scene}}_{j}}\right\|}, \\
    &\ q_{i}^{align}=\frac{e^{s_{i}^{scene}}}{\sum_{j=1}^{N}e^{s_{j}^{scene}}}, i,j \in\{1,2,\dots,N\}.
\end{aligned}
\end{equation}
To ensure full alignment of the image with the high-level scene semantic information, we set the ground truth alignment levels $g^{align}$ to be 1. This follows the image-text alignment dataset~\cite{10262331}, where the alignment score is scaled from 0 to 1. Thus, we set the maximum alignment level as 1.
\begin{equation}
\small
L_{A} =\frac{1}{N} \sum_{i=1}^{N}\left\|q^{align}_{i}-g^{align}_{i}\right\|_{2}^{2},
\end{equation}

The above loss calculation assumes that the textual descriptions generated by InstructBLIP perfectly align with the corresponding images. The final loss is calculated as follows:
\begin{equation}
\small
L = \alpha L_A +\beta L_C,
\label{eqhyper}
\end{equation}
where $\alpha$ and $\beta$ are the weight hyperparameters that determine the proportion of two branches, which are empirically set to 0.1 and 0.9, respectively.
\section{Experiments}
\subsection{Dataset and Evaluation Metric}
We conduct experiments on three publicly available image complexity assessment datasets: IC9600~\cite{9999482}, VISC-C~\cite{kyle2023characterising}, and Savoias~\cite{saraee2018savoias}. IC9600 is the first large-scale ICA dataset, containing 9,600 images across eight semantic categories, each annotated by 17 trained annotators. VISC-C includes 800 images with corresponding human-rated complexity scores. Savoias comprises over 1,400 images from seven categories, covering a broad range of low- and high-level visual features.

Following ICNet~\cite{9999482}, we evaluate model performance using four metrics: Pearson Linear Correlation Coefficient (PLCC), Spearman Rank-Order Correlation Coefficient (SRCC), Root Mean Squared Error (RMSE), and Root Mean Absolute Error (RMAE). PLCC and SRCC assess correlation with subjective scores, while RMSE and RMAE quantify prediction errors. Together, these metrics offer a comprehensive evaluation of prediction accuracy and consistency.
\begin{table}[t]
\centering
\footnotesize
\caption{Performance comparisons on the IC9600, VISC-C, and Savoias datasets. The best-performing results are highlighted in bold. The calculation of model Flops is based on the input image shape $3\times512\times512$, and the inference speed is taken from an average of 100 inferences. * The CLICv2 is pretrained on IC1M dataset and obtained by linear probing on the IC9600. * The Savoias results of DINO-v2 are pretrained on IC9600.}
\resizebox{\linewidth}{!}{
\begin{tabular}{c|c|c|c|c|c|c|c|c|c|c|c|c}
\hline
Dataset & \multicolumn{3}{c|}{IC9600} & \multicolumn{3}{c|}{VISC-C} &\multicolumn{3}{c|}{Savoias} &\multicolumn{3}{c}{Complexity}\\
\hline
Model & SRCC$\uparrow$ & PLCC$\uparrow$ & RMSE$\downarrow$ & SRCC$\uparrow$ & PLCC$\uparrow$ & RMSE$\downarrow$ & SRCC$\uparrow$ & PLCC$\uparrow$ & RMSE$\downarrow$ & Params/M & Flops/G & Speed/ms\\
\hline
CR~\cite{Machado_Romero_Nadal_Santos_Correia_Carballal_2015} & 0.314 & 0.228 & 0.196 & - & - & - & 0.305 & 0.271 & 0.257 & - & - & -\\
ED~\cite{GUO2018110} & 0.491 & 0.569 & 0.226 & - & - & - & 0.449 & 0.467 & 0.273 & - & - & -\\
DBCNN~\cite{8576582} & 0.871 & 0.879 & 0.071 & 0.779 & 0.783 & 0.086 & 0.768 & 0.770 & 0.147 & 15.31 & 86.22 & 23.69\\
NIMA~\cite{talebi2018nima} & 0.838 & 0.555 & 0.194 & 0.810 & 0.803 & 0.125 & 0.781 & 0.771 & 0.210 & 54.32 & 13.18 & 25.00\\
HyperIQA~\cite{Su_2020_CVPR} & 0.926 & 0.933 & 0.204  & 0.734 & 0.739 & 0.181 & 0.801 & 0.798 & 0.293 & 27.38 & 107.83 & 29.64\\
CLIPIQA~\cite{wang2023exploring} & 0.897 & 0.898 & 0.078 & 0.781 & 0.796 & 0.122 & 0.779 & 0.794 & 0.101 & - & 61.07 & 21.76\\
TOPIQ~\cite{chen2024topiq} & 0.938 & 0.944 & 0.049 & 0.803 & 0.811 & 0.079 & 0.838 & 0.832 & 0.123 & 45.20 & 37.26 & 14.43\\
CNet~\cite{kyle2022predicting} & 0.870 & 0.873 & -  & - & 0.716 & - & - & - & - & - & - & -\\
*CLICv2~\cite{liu2025clicv2} & 0.927 & 0.933 & -  & - & & - & - & - & - & - & - & -\\
*DINO-v2~\cite{celona2024use} & 0.847 & 0.851 & 0.086 & - &  & - & 0.675 & 0.689 & - & - & - & -\\
ICCORN~\cite{guo2023image} & 0.951 & 0.954 & 0.048  & 0.766 & 0.784 & 0.332 & - & - & - & - & - & -\\
ICNet~\cite{9999482} & 0.937 & 0.946 & 0.049 & 0.818 & \textbf{0.814} & 0.079 & 0.865 & 0.849 & 0.121 & 20.33 & 28.40 & 4.56\\
\textbf{CM-SSA} & \textbf{0.958} & \textbf{0.961} & \textbf{0.009} & \textbf{0.823} & 0.805 & \textbf{0.018} & \textbf{0.883} & \textbf{0.875} & \textbf{0.026} & 205.45 & 52.66 & 35.43 \\
\hline
\end{tabular}}
\label{comparison}
\end{table}
\subsection{Implement Details}
We adopt a CLIP model based on ViT-B/32, setting the length of the learnable prompts to 8. The datasets are randomly divided into training and testing sets with an 8:2 ratio. During training, the batch size per iteration is 64, and the model is optimized using the Adam algorithm with a learning rate of $10^{-4}$ over 50 epochs. In the testing phase, complexity scores are predicted using a patch-based evaluation method~\cite{fu2024vision}. All experiments are implemented in PyTorch on a V100 GPU.
\subsection{Performance Comparisons}
Following the protocols of the ICNet~\cite{9999482}, we conduct a performance comparison of the proposed CM-SSA with two hand-crafted feature-based ICA methods, two deep learning-based ICA methods, and five advanced image quality assessment (IQA) models: The ICA baselines include compression ratio (CR) ~\cite{Machado_Romero_Nadal_Santos_Correia_Carballal_2015},  edge density (ED)~\cite{GUO2018110}, CNet~\cite{kyle2022predicting}, CLICv2~\cite{liu2025clicv2}, DINO-v2~\cite{celona2024use}, ICCORN~\cite{guo2023image}, and ICNet~\cite{9999482}. The IQA baselines include DBCNN~\cite{8576582}, NIMA~\cite{talebi2018nima}, HyperIQA~\cite{Su_2020_CVPR}, CLIPIQA~\cite{wang2023exploring}, and TOPIQ~\cite{chen2024topiq}.

As shown in Table~\ref{comparison}, hand-crafted feature-based ICA methods (CR, ED) underperform significantly across all datasets due to their limited capacity to model high-level semantics. While several IQA methods (e.g., DBCNN, NIMA, and HyperIQA) show moderate performance, they fall short on ICA tasks as they are not specifically designed for complexity perception. In contrast, deep ICA models (ICCORN, ICNet, and CM-SSA) achieve consistently superior results, highlighting the importance of task-specific learning. Notably, our proposed CM-SSA outperforms all other methods across datasets, demonstrating state-of-the-art performance with competitive computational efficiency.

\subsection{Cross-Dataset Evaluation}
Table \ref{crossdata} presents cross-dataset validation results, evaluating the generalization capability of various models when trained and tested on different datasets. The proposed model consistently demonstrates stronger robustness across most cross-dataset scenarios. For example, when trained on IC9600 and tested on VISC-C, our model achieves the highest SRCC and PLCC values of 0.760 and 0.740, respectively. However, CM-SSA exhibits reduced generalization performance when trained on VISC-C and tested on other datasets. This may be due to the limited scale and diversity of the VISC-C dataset, especially given the relatively large number of parameters in CM-SSA that make it more susceptible to generalizing from a small dataset to larger ones. These results confirm that the proposed method generalizes competitively across diverse datasets with varying characteristics.
\begin{table}[t]
\centering
\footnotesize
\caption{Cross-dataset validation. The best performances are highlighted in bold.}
\resizebox{\linewidth}{!}{
\begin{tabular}{c|cccc|cccc|cccc}
\hline
Train & \multicolumn{4}{c|}{IC9600} & \multicolumn{4}{c|}{VISC-C} & \multicolumn{4}{c}{Savoias} \\
\hline
Test & \multicolumn{2}{c}{VISC-C} & \multicolumn{2}{c|}{Savoias} & \multicolumn{2}{c}{IC9600} & \multicolumn{2}{c|}{Savoias} & \multicolumn{2}{c}{IC9600} & \multicolumn{2}{c}{VISC-C} \\
\hline
Model & \multicolumn{1}{l}{SRCC} & \multicolumn{1}{l}{PLCC} & \multicolumn{1}{l}{SRCC} & \multicolumn{1}{l|}{PLCC} & \multicolumn{1}{l}{SRCC} & \multicolumn{1}{l}{PLCC} & \multicolumn{1}{l}{SRCC} & \multicolumn{1}{l|}{PLCC} & \multicolumn{1}{l}{SRCC} & \multicolumn{1}{l}{PLCC} & \multicolumn{1}{l}{SRCC} & \multicolumn{1}{l}{PLCC} \\
\hline
DBCNN & 0.685 & 0.674 & 0.606 & 0.626 & 0.742 & 0.750 & 0.525 & 0.538 & 0.773 & 0.787 & 0.644 & 0.652 \\
NIMA & 0.566 & 0.297 & 0.603 & 0.240 & 0.714 & 0.390 & 0.575 & 0.540 & 0.818 & 0.725 & 0.630 & 0.588 \\
HyperIQA & 0.711 & 0.688 & 0.669 & 0.669 & 0.668 & 0.668 & 0.550 & 0.554 & 0.761 & 0.748 & 0.643 & 0.629 \\
CLIPIQA & 0.702 & 0.686 & 0.600 & 0.608 & 0.406 & 0.399 & 0.487 & 0.481 & 0.789 & 0.775 & 0.550 & 0.568 \\
TOPIQ & 0.759 & 0.730 & 0.659 & 0.648 & 0.695 & 0.700 & 0.594 & 0.597 & 0.806 & 0.804 & 0.674 & 0.685 \\
ICNet & 0.753 & 0.712 & \textbf{0.728} & \textbf{0.716} & \textbf{0.718} & \textbf{0.739} & \textbf{0.664} & \textbf{0.660} & 0.805 & 0.807 & \textbf{0.712} & \textbf{0.709} \\
\textbf{CM-SSA} & \textbf{0.760} & \textbf{0.740} & 0.710 & 0.712 & 0.577 & 0.556 & 0.541 & 0.486 & \textbf{0.838} & \textbf{0.828} & 0.702 & 0.681 \\
 \hline
\end{tabular}}
\label{crossdata}
\end{table}
\begin{table}[t]
\centering
\begin{minipage}{0.45\linewidth}
\caption{Performance comparison of different branch configurations. The text prompts are coarse-grained complexity prompts when the single complexity regression branch $\rm {Branch}_{C}$ is utilized, and the text prompts are fine-grained textual prompts when the single text-driven scene semantic branch $\rm {Branch_{A}}$ is utilized. The best results are highlighted in bold.}
\resizebox{\linewidth}{!}{
\begin{tabular}{cc|cccc}
\hline
$\rm {Branch}_{C}$   & $\rm {Branch_{A}}$   &   SRCC  $\uparrow$   & PLCC $\uparrow$   &  RMSE$\downarrow$ &   RMAE$\downarrow$ \\ \hline
\checkmark & $\times$ & 0.947 & 0.951 & 0.010 & 0.008 \\
$\times$ & \checkmark & 0.915 & 0.922 & 0.013 & 0.010 \\
\checkmark & \checkmark & \textbf{0.958} & \textbf{0.961} & \textbf{0.009} & \textbf{0.007} \\ \hline
\end{tabular}}
\label{tab:branch_performance}
\end{minipage}
\hfill
\begin{minipage}{0.45\linewidth}
\centering
\caption{Ablation study on the prompt levels in $\rm {Branch_{C}}$. 3 levels refer to [``Simple Complexity, Moderate Complexity, High Complexity''], while 7 levels refer to [``Very Low Complexity, Fairly Low Complexity, Slightly Low Complexity, Moderate Level Complexity, Slightly High Complexity, Fairly High Complexity, Very High Complexity''].}
\resizebox{\linewidth}{!}{
\begin{tabular}{c|cccc}
\hline
Class Name & SRCC $\uparrow$ & PLCC$\uparrow$ & RMSE$\downarrow$ & RMAE $\downarrow$ \\
\hline
3 levels & 0.951 & 0.954 & 0.010 & 0.008 \\
5 levels & \textbf{0.958} & \textbf{0.961} & \textbf{0.009} & \textbf{0.007} \\
7 levels & 0.955 & 0.960 & \textbf{0.009} & \textbf{0.007}\\
\hline
\end{tabular}}
\label{tab:ablationlevel}
\end{minipage}
\end{table}
\begin{table}[t]
\centering
\begin{minipage}{0.45\linewidth}
\centering
\caption{Ablation study on the prompt models and lengths on IC9600 dataset.}
\resizebox{\linewidth}{!}{
\begin{tabular}{c|cccc}
\hline
Prompt & SRCC $\uparrow$ & PLCC$\uparrow$ & RMSE$\downarrow$ & RMAE $\downarrow$\\
\hline
BLIP & 0.947 & 0.955 & \textbf{0.009} & \textbf{0.007} \\
BLIP2 & 0.952 & 0.957 & 0.010 & \textbf{0.007} \\
InstructBLIP-Short & 0.952 & 0.956 & 0.010 & 0.008 \\
InstructBLIP-Medium & \textbf{0.958} & \textbf{0.961} & \textbf{0.009} & \textbf{0.007} \\
InstructBLIP-Long & 0.953 & 0.957 & 0.010 & 0.008\\
\hline
\end{tabular}}
\label{tab:ablationprompt}
\end{minipage}
\hfill
\begin{minipage}{0.45\linewidth}
\centering
\caption{Ablation study on the hyper-parameters $\alpha$ and $\beta$ on IC9600 dataset.}
\resizebox{\linewidth}{!}{
\begin{tabular}{cc|cccc}
\hline
$\alpha$ & $\beta$ & SRCC $\uparrow$ & PLCC $\uparrow$ & RMSE $\downarrow$ & RMAE $\downarrow$ \\
\hline
0.1 & 0.9 & \textbf{0.958} & \textbf{0.961} & \textbf{0.009} & \textbf{0.007} \\
0.3 & 0.7 & 0.952 & 0.956 & 0.010 & 0.008 \\
0.5 & 0.5 & 0.952 & 0.955 & 0.010 & 0.008 \\
0.7 & 0.3 & 0.951 & 0.954 & 0.010 & 0.008 \\
0.9 & 0.1 & 0.950 & 0.954 & 0.010 & 0.008\\
\hline
\end{tabular}}
\label{tab:ablationhyper}
\end{minipage}
\end{table}

\subsection{Ablation Tests}
\label{branches}
\begin{itemize}
    \item \textbf{Validity on branches}. Table \ref{tab:branch_performance} presents a performance comparison between the two branch configurations for image complexity assessment. Using only the complexity regression branch yields relatively high performance, indicating its significant role. In contrast, relying solely on the scene semantic alignment branch results in lower performance, highlighting limited standalone effectiveness. Meanwhile, combining both branches yields the best results, emphasizing the complementary nature of the two branches. This confirms that although scene semantic information is less effective for direct complexity regression, it plays a crucial role in refining image features to be more perceptually relevant to scene semantic information.
    \item \textbf{Validity on prompt level in $\rm Branch_C$}. Table \ref{tab:ablationlevel} displays the ablation on prompt levels. The results reveal that a finer level of text prompt in $\rm Branch_C$ can lead to better complexity prediction.
    \item \textbf{Validity on prompt length in $\rm Branch_A$}. Table \ref{tab:ablationprompt} compares different prompt lengths and models. InstructBLIP outperforms BLIP and BLIP2, likely because BLIP-based models generate shorter prompts. Regarding prompt length, both short and long prompts underperform the medium ones, possibly due to insufficient information in short prompts and hallucinations in long prompts.
    \item \textbf{Validity on hyperparameters}. Table \ref{tab:ablationhyper} presents the ablation results on the hyperparameters (i.e., $\alpha, \beta$ in eq. \ref{eqhyper}). The results illustrate that the $\rm Branch_C$ contributed most to the final complexity regression, while increasing the proportion of $\alpha$ may result in a slight performance drop.
\end{itemize}

\section{Conclusion}
In this paper, we present the Cross-Modal Scene Semantic Alignment (CM-SSA), the first cross-modal ICA framework. The proposed CM-SSA addresses the shortcomings of existing ICA metrics, which largely depend on hand-crafted and deep learning-based features limited to single-modality visual information. Specifically, our proposed CM-SSA is a dual-branch architecture. The complexity regression branch leverages pair-wise learning between images and their corresponding complexity level prompts to predict complexity scores. To further enhance the model's understanding of scene semantic information, the scene semantic alignment branch is designed to align image features with scene semantic descriptions. This alignment refines visual prompts with high-level text-driven scene semantics, enriching the complexity assessment process. Experimental results on several ICA datasets demonstrate that the proposed CM-SSA outperforms state-of-the-art methods. Ablation studies further validate the contributions of each branch and the generalization performance across datasets.

\small
\bibliography{egbib}
\end{document}